\documentclass[10pt,twocolumn,letterpaper]{article}

\usepackage{cvpr}
\usepackage{times}
\usepackage{epsfig}
\usepackage{graphicx}
\usepackage{amsmath}
\usepackage{amssymb}
\usepackage[linesnumbered,ruled,vlined]{algorithm2e}
\usepackage{floatrow}
\usepackage[font=footnotesize,margin=2pt,skip=2pt]{caption}
\usepackage[breaklinks=true,bookmarks=false]{hyperref}
\usepackage{caption}

\cvprfinalcopy  %*** Uncomment this line for the final submission

\def\cvprPaperID{2971} % *** Enter the CVPR Paper ID here

\ifcvprfinal\fi
\begin{document}

%%%%%%%%% TITLE
% \title{Efficient Multi-Domain Network Learning by Covariance Normalization}
\title{Efficient Multi-Domain Learning by Covariance Normalization}

\author{Yunsheng Li \quad \quad Nuno Vasconcelos\\
University of California San Diego\\
La Jolla, CA 92093\\
{\tt\small yul554@ucsd.edu, nvasconcelos@ucsd.edu}
%\and
%Nuno Vasconcelos\\
%UC San Diego\\
}

\maketitle
\thispagestyle{empty}

%%%%%%%%% ABSTRACT
\begin{abstract}
   The problem of multi-domain learning of deep networks is considered.
  An adaptive layer is induced per target domain and a novel
  procedure, denoted {\it covariance normalization\/} (CovNorm),
  proposed to reduce its parameters. CovNorm is a data driven
  method of fairly simple implementation, requiring two principal component
  analyzes (PCA) and fine-tuning of a mini-adaptation layer.
  Nevertheless, it is shown, both theoretically and experimentally,
  to have several advantages over previous approaches, such
  as batch normalization or geometric matrix approximations.
  Furthermore, CovNorm can be deployed both when target datasets are
  available sequentially or simultaneously. Experiments show that, in both cases, it has performance
  comparable to a fully fine-tuned network, using as few as
  $0.13\%$ of the corresponding parameters per target domain.
\end{abstract}

\section{Introduction}

\begin{figure}[t]\RawFloats
  \centering
  \includegraphics[width=3in]{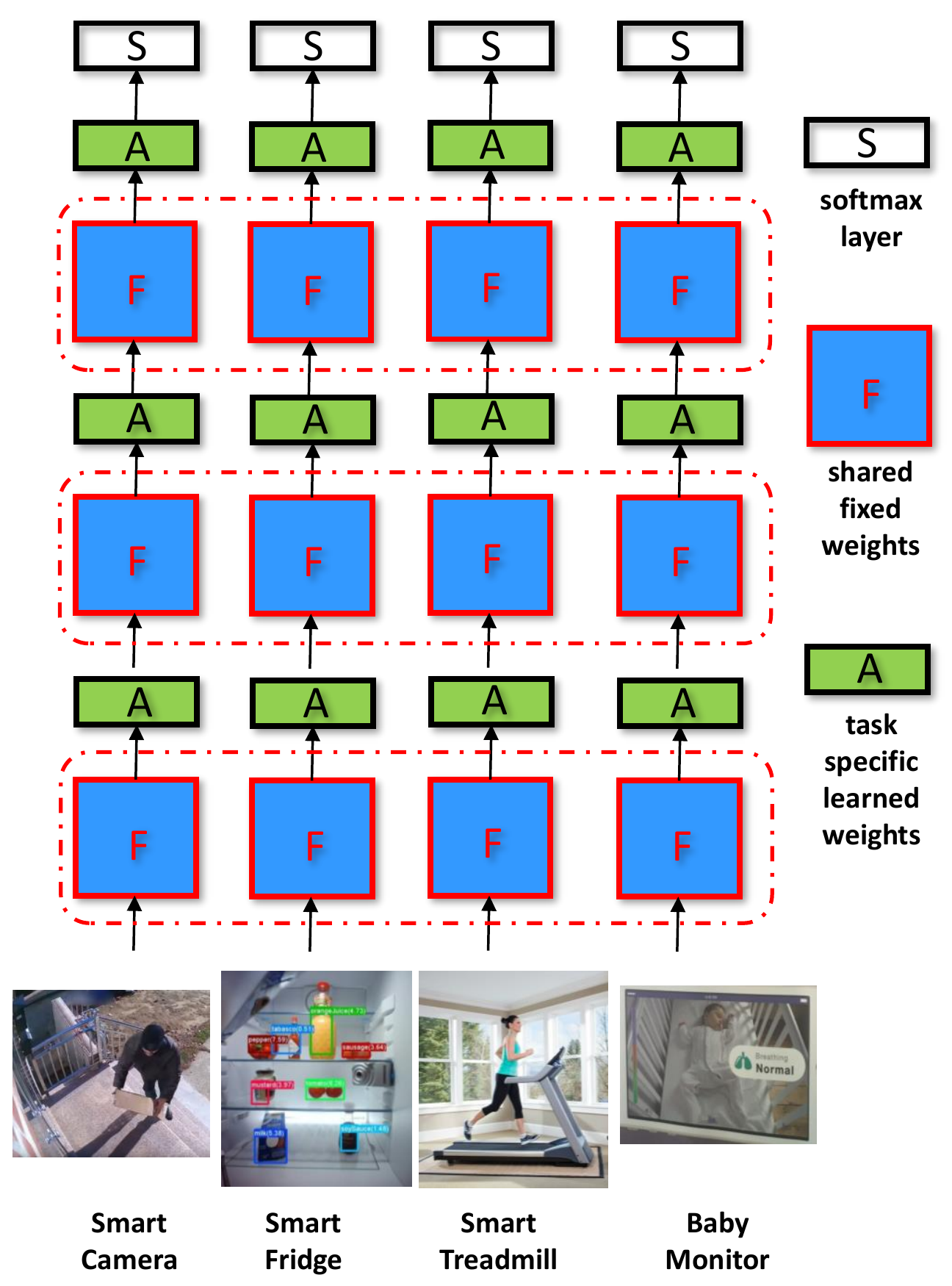}
  \caption{Multi-domain learning addresses the efficient 
    solution of several tasks, defined on different domains. 
    Each task is solved by a different network but all networks share a set of
    fixed layers $\bf F$, which contain the majority of network
    parameters. These are complemented by small task-specific
    adaptation layers $\bf A$.}
  \label{fig:BlockD}
\end{figure}

Convolutional nerual networks (CNNs) have enabled transformational advances
in classification, object detection and segmentation, among other tasks.
However they have non-trivial complexity. State of the art models contain 
millions of parameters and require
implementation in expensive GPUs. This creates problems for applications
with computational constraints, such as mobile devices or consumer
electronics. Figure~\ref{fig:BlockD} illustrates the problem in the
context of a smart home equipped with an ecology of devices such as a 
camera that monitors package delivery and theft, a fridge that keeps
track of its content, a treadmill that adjusts fitness routines
to the facial expression of the user, or a baby monitor that keeps
track of the state of a baby. As devices are added to the ecology, the
GPU server in the house must switch between a larger number of classification,
detection, and segmentation tasks. Similar problems will be faced by
mobile devices, robots, smart cars, etc.

Under the current deep learning paradigm, this task
switching is difficult to perform. The predominant strategy is to use
a different CNN to solve each task.
%usally fine-tuned from a model
%trained on a large dataset, e.g. the VGG or ResNet networks trained on
%ImageNet.
Since only a few models can be cached in the GPU, and
moving models in and out of cache adds too much overhead to enable
real-time task switching, there is a need for very efficient
parameter sharing across tasks. The individual networks should 
share most of their parameters, which would always reside on the GPU. 
A remaining small number of task specific parameters would be switched 
per task. This problem is known as {\it multi-domain
  learning\/} (MDL) and has been addressed with the architecture of
Figure~\ref{fig:BlockD}~\cite{rebuffi2017learning,rosenfeld2017incremental}. This consists
of set of  {\it fixed\/} layers (denoted as '$\bf{F}$') shared
by all tasks and a set of task specific {\it adaptation\/} layers (denoted as 
'$\bf{A}$') fine-tunned to each task. If the $\bf A$ layers are much
smaller than the $\bf F$ layers, many models can be cached simultaneously.
Ideally, the $\bf F$ layers should
be pre-trained, e.g. on ImageNet, and used by all tasks without
additional training, enabling the use
of special purpose chips to implement the majority of the computations.
While $\bf A$ layers would still require a processing
unit, the small amount of computation could enable the use
of a CPU, making it cost-effective to implement each 
network on the device itself.

% %We believe that multi-domain learning is a misnomer for this problem and, for
% %reasons discussed in Section~\ref{sec:related}, and it
% %referred to as {\it transfer learning with architectural constraints\/}.
% Under this view, a fixed network $\bf F$ is {\it transferred\/} to an ecology
% of tasks by the addition (and finetunning) of task specific $\bf T$ layers.
% The original $\bf F$ layers are {\it constrained\/} to remain unaltered
% and are shared by all tasks.
% %However, for consistency
% %with~\cite{rebuffi17learning,rosenfeld17}, we
% %will continue to refer to it as ``multidomain learning.''
% % Due to this, TLAC has aspects in common with ``parameter sharing'' 
% % problems such as transfer learning~\cite{...}, 
% % domain adaptation~\cite{...}, multitask learning~\cite{...}, multidomain
% % learning~\cite{...}, and life-long learning~\cite{...}. 
% % For example, the ecology solves multiple tasks, whose inputs can be of 
% % the same or different domains, and networks may be learned simultaneously or 
% % added to the ecology incrementally. However, TLAC is not an instance 
% % of any of these problems. For example, it is not a strict
% % multi-task learning problem because devices are not allways used 
% % simultaneously. 

\begin{figure}[t]
  \centering
  \includegraphics[width=.9\linewidth]{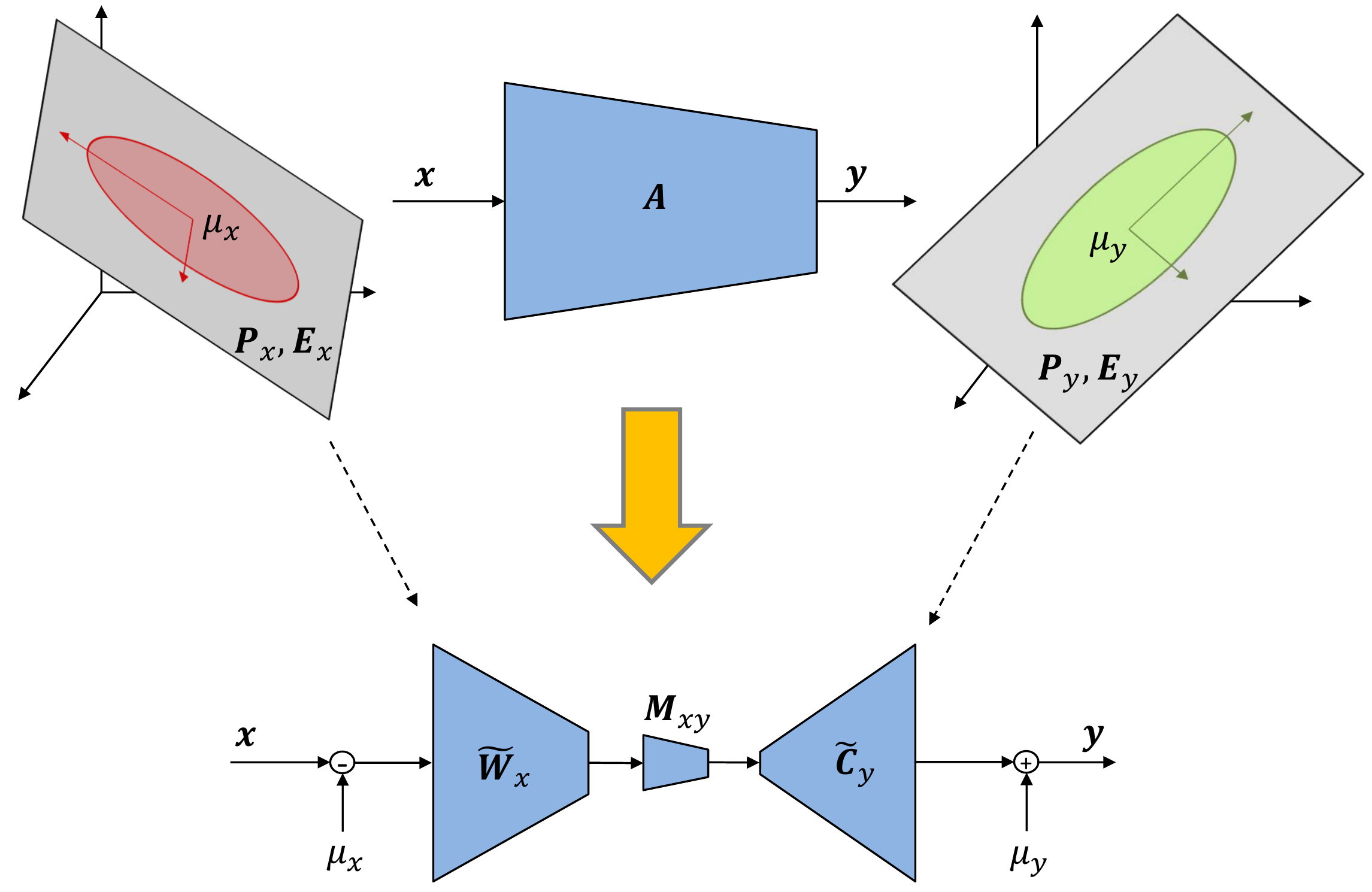}
  \caption{Covariance normalization. Each adaptation layer ${\bf A}$ is
    approximated by three transformations: $\tilde{\bf W}_x$,
    which implements a projection onto the PCA space of the input ${\bf x}$
    (principal component matrix ${\bf P}_x$ and eigenvalue matrix ${\bf E}_x$),
    $\tilde{\bf W}_y$, which reconstructs the PCA space of
    the output $\bf y$ (matrices ${\bf P}_y$ and ${\bf E}_y$), and
    a mini-adaptation layer ${\bf M}_{xy}$.}
  \label{fig:covnorm}
\end{figure}

In summary, MDL aims to maximize the performance of the network
ecology  while minimizing the ratio of task specific ($\bf A$) to
total parameters (both types ${\bf F}$ and ${\bf A}$) per network.
\cite{rebuffi2017learning,rosenfeld2017incremental} have shown 
%using the well known ResNet residual
%structure~\cite{he2016deep} to implement the $\bf T$ layers, which were
%denoted {\it residual adaptation\/} (RA) layers. , they showed
that the architecture of Figure~\ref{fig:BlockD} can match the performance
of fully fine-tuning each network in the ecology, even when $\bf A$ layers
contain as few as $10\%$ of the 
total parameters.
%Nevertheless, this can still be a large number of
%parameters for many applications.
In this work, we show that $\bf A$ layers can be substantially further shrunk,
% focus on the first problem in MDL in which case the datasets can only be accessed incrementally.
%We start with the RA layers of \cite{rebuffi2017learning} and
using a data-driven  low-rank approximation. As illustrated in
Figure~\ref{fig:covnorm}, this is based on transformations
that match the $2^{nd}$-order statistics of the $\bf{A}$ layer inputs and
outputs. Given principal component analyses (PCAs) of
both input and output, the layer is approximated by a
{\it recoloring transformation:\/} a projection into input PCA space, 
followed by a reconstruction into the output PCA 
space. By controlling the intermediate PCA dimensions, the method enables
low-dimensional approximations of different input and output dimensions. 
To correct the mismatch (between PCA components) of two PCAs learned 
independently, a small {\it mini-adaptation\/} layer is introduced 
between the two PCA matrices, and fine-tunned on the target target.

Since the overall transformation generalizes {\it batch normalization\/},
the method is
denoted {\it covariance normalization\/} (CovNorm). CovNorm is shown to
outperform, with both theoretical and experimental arguments, purely
geometric methods for matrix approximation, such as the singular value
decomposition (SVD)~\cite{rebuffi2018efficient}, fine-tuning of the original $\bf A$ 
layers~\cite{rebuffi2017learning,rosenfeld2017incremental}, or adaptation based on batch 
normalization~\cite{bilen2017universal}. It is also quite simple, requiring two PCAs and 
the finetuning of a very small mini-adaptation layer per $\bf{A}$ layer and 
task. Experimental results show that it can outperform full network 
fine-tuning while reducing $\bf{A}$ layers to
as little as $0.53\%$ of the total parameters. When all tasks can be
learned together, $\bf{A}$ layers can be further reduced to $0.51\%$ of the
full model size. This is achieved by combining the individual PCAs into a
global PCA model, of parameters shared by all tasks, and only fine-tunning
mini-adaptation layers in a task specific manner.

\section{Related work}

MDL is a transfer learning problem, namely the transfer of a
model trained on a {\it source\/} learning problem to an ecology
of {\it target\/} problems. This makes it related to different types of
transfer learning problems, which differ mostly in terms of input, or
{\it domain,\/} and range space, or {\it task\/}. 

{\bf Task transfer:} Task transfer addresses the use of a model trained on a 
source task to the solution of a target task. The two tasks can be defined 
on the same or different domains. Task transfer is prevalent in deep learning,
where a CNN pre-trained on a large source dataset, such as ImageNet, is usually 
fine-tunned~\cite{lecun2015deep} to a target task. While extremely effective and 
popular, full network fine-tunning changes most network parameters,
frequently all. MDL addresses this problem by considering multiple target
tasks and extensive parameter sharing between them.

{\bf Domain Adaptation:} In domain adaptation, the source and
target tasks are the same, and a model trained on a source domain is 
transfered to a target domain. Domain adaptation
can be supervised, in which case labeled data is available
for the target domain, or unsupervised, where it is not. Various
strategies have been used to address these problems. Some methods
seek the network parameters that minimize some function of the distance
between feature distributions in the two domains~\cite{long2015learning,bousmalis2016domain,sun2016deep}. Others
introduce an adversarial loss that maximizes the confusion between the two 
domains~\cite{ganin2014unsupervised,tzeng2017adversarial}. A few methods have also proposed to do the transfer at 
the image level, e.g. using GANs~\cite{goodfellow2014generative} to map 
source images into (labeled) target images, then used to learn a target 
classifier~\cite{bousmalis2017unsupervised,shrivastava2017learning,
  hoffman2017cycada}. All these methods exploit the commonality of 
source and domain tasks to align source and target domains.
This is unlike MDL, where source and target tasks are
different. Nevertheless, some mechanisms proposed for domain
adaptation can be used for MDL. For example,
\cite{carlucci2017autodial,mancini2018boosting} use a batch normalization layer 
to match the statistics of source and target data, in terms of means and 
standard deviation. This is similar to an early proposal for
MDL~\cite{bilen2017universal}. We show that these mechanisms underperform covariance
normalization. 

% Then adversarial loss \cite{goodfellow2014generative} was used to reduce the distribution gap between the source and the target dataset. In \cite{tzeng2017adversarial}, they add a discriminator on top of the last layers of CNN and train the network in an adversarial manner and get a good result for digit datasets. The difference between our work and DA is in our work we don't require the datasets share same content and instead of totally missing ground truth for the target dataset we can access the labels for the dataset we want to learn. 

\begin{figure*}[t]\RawFloats
  \centering
  \begin{tabular}{ccc}
    \includegraphics[width=.28\linewidth]{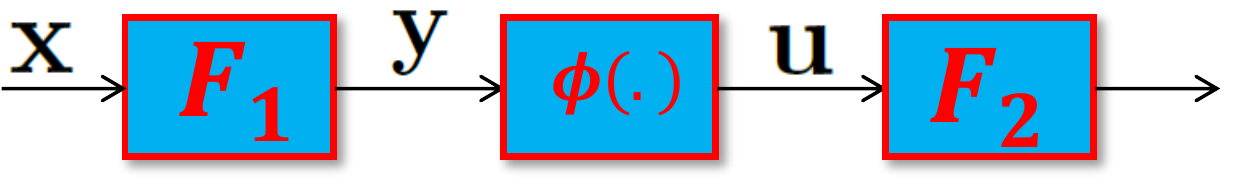} 
    & \includegraphics[width=.28\linewidth]{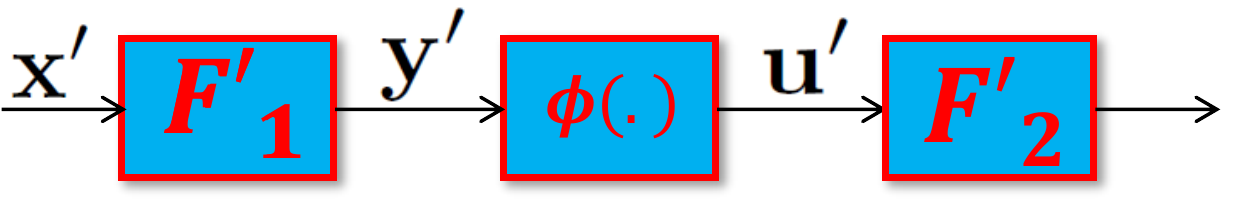} 
    & \includegraphics[width=.33\linewidth]{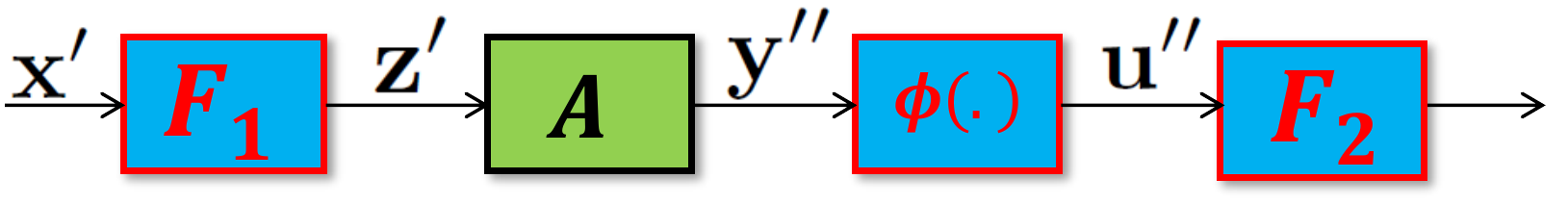}\\
    \footnotesize{a)} & \footnotesize{b) } & \footnotesize{c) }\\
  \end{tabular}
  \caption{a) original network, b) after fine-tuning, and c) with adaptation
    layer $\bf A$. In all cases, ${\bf W}_i$ is a weight layer and
  $\phi(.)$ a non-linearity.}
  \label{fig:intro}
\end{figure*}

{\bf Multitask learning:} Multi-task
learning~\cite{caruana1998multitask,zhang2017survey}
addresses the solution of multiple tasks by the same model.
It assumes that all tasks have the same visual domain. Popular examples
include classification and bounding box regression in
object detection \cite{girshick2015fast,ren2017faster}, joint estimation
of surface normals and depth~\cite{eigen2015predicting} or
segmentation~\cite{MisraCrossMTL16}, joint representation in terms of
attributes and facial landmarks~\cite{zhang2014facial,ranjan2017hyperface}, among
others. Multitask learning is sometimes also used to solve
auxiliary tasks that strengthen performance of a task of interest, e.g. by
accounting for context~\cite{gkioxari2015contextual}, or representing objects
in terms of classes and
attributes~\cite{huang2015cross,MisraCrossMTL16,morgado2017semantically,
  lu2017fully}.
Recently, there have been attempts to learn models that solve many
problems
jointly~\cite{kendall2017multi,kokkinos2017ubernet,zamir2018taskonomy}.
% For example,
% \cite{kendall2017multi} proposed to jointly learn pixel depth regression,
% semantic and instance segmentation from a monocular image,
% while~\cite{kokkinos2017ubernet} introduced a network for joint estimation
% of boundaries, saliency, normals, segmentation, human parts and object
% bounding boxes and \cite{zamir2018taskonomy} studied the relationships
% between many tasks in the visual domain.

Most multitask learning
approaches emphasize the learning of the interrelationships between
tasks. This is frequently accomplished by using a single network,
combining domain agnostic lower-level network layers with task specific
network heads and loss
functions~\cite{zhang2014facial,eigen2015predicting,gkioxari2015contextual,huang2015cross,ren2017faster,kokkinos2017ubernet}, or some more sophisticated forms of
network branching~\cite{lu2017fully}. The branching architecture is
incompatible with MDL, where each task has its own input,
different from those of all other tasks. Even when multi-task learning is
addressed with multiple tower networks, the emphasis tends to be on 
inter-tower connections, e.g. through
cross-stitching~\cite{MisraCrossMTL16,jou2016deep}. In MDL, such connections
are not feasible, because different networks can join the ecology of
Figure~\ref{fig:BlockD} asynchronously, as devices are turned on and
off.

% solve other problems, e.g. semantic
% segmentation~\cite{long2015fully} and natural language
% processing \cite{collobert2008unified}, from a collection of
% datasets. For example,  
% %They weight different tasks based on the homoscedastic uncertainty and in 

% A common assumption in multi-task learning is that
% a single model solves many tasks simultaneously. These tasks are usually
% defined on a single domain, e.g. segmentation and depth estimation of
% a single image. These assumptions are not valid for MTWCL, where each
% task can have a different domain and the tasks do not have to be solved
% simultaneously.

% Differetn from our work, most of the MTL papers try to solve multiple problems on the same dataset which means the domain is the same but all the tasks can share one model. Following \cite{rebuffi2017learning}, in our work, we try to solve the classification problem on multiple visual domain which is referred as MDL. 

{\bf Lifelong learning:} Lifelong learning aims to learn multiple tasks
sequentially with a shared model. This can be done by adapting the
parameters of a network or adapting the network architecture.
Since training data is discarded upon its use, constraints are needed
to force the model to remember what was previously learned.
Methods that only change parameters either use the model output on previous
tasks \cite{li2017learning}, previous parameters
values \cite{lee2017overcoming}, or previous network
activations \cite{triki2017encoder} to regularize the learning of the
target task. They are very effective at parameter sharing, since a
single model solves all tasks. However, this model is not optimal for
any specific task, and can perform poorly on all tasks, depending on
the mismatch between source and target domains~\cite{rebuffi2017icarl}.
We show that they can significantly underperform MDL with CovNorm.
Methods that adapt the network architecture
usually add a tower per new task~\cite{rusu2016progressive,aljundi2017expert}.
These methods have much larger complexity than MDL, since several
towers can be needed to solve a single task~\cite{rusu2016progressive},
and there is no sharing of fixed layers across tasks.
% Recent works address this in two ways: extra memory or extra parameters. \cite{lopez2017gradient} introduced an ``episodic memory'' that maintains a subset of the samples from previous tasks.
% Given a new target, the CNN is updated with a gradient computed from both new and old samples.

 %Our work is inspired by this method and uses the same RA architecture, investigating further reductions of number of RA parameters per task. We show that both RA and FNFT can be outperformed with as few as $0.4\%$ new parameters per dataset. 

{\bf Multi-domain learning:} This work builds on previous
attempts at MDL, which have investigated different architectures for
the adaptation layers of Figure~\ref{fig:BlockD}.
\cite{bilen2017universal} used a BN layer \cite{ioffe2015batch} of
parameters tunned per task. While performing well on simple datasets,
this does not have enough degrees of freedom to support transfer of large
CNNs across different domains. More powerful architectures were
proposed by \cite{rosenfeld2017incremental}, who used a $1 \times 1$
convolutional layer
and \cite{rebuffi2017learning}, who proposed a
ResNet-style residual layer, known as a residual adaptation (RA) module.
These methods were shown to perform surprisingly well in terms of recognition
accuracy, equaling or surpassing the performance of full network fine
tunning, but can still require a substantial number of adaptation
parameters, typically $10\%$ of the network size.
\cite{rebuffi2018efficient} addressed this problem by combining
adapters of multiple tasks into a large matrix, which is approximated with
an SVD. This is then fine-tuned on each target dataset. Compressing
adaptation layers in this way was shown to reduce adaptive parameter counts to
approximately half of \cite{rebuffi2017learning}. However, all tasks
have to be optimized simultaneously. We show that CovNorm enables
a further ten-fold reduction in adaptation layer parameters,
without this limitation, although some additional gains are possible
with joint optimization.

%\cite{kaiser2017one} proposed
%a single model that can solve $8$ tasks, ranging from ImageNet
%classification \cite{krizhevsky2009learning} to image
%captioning \cite{lin2014microsoft} and language
%translation \cite{kim2017joint}.

%We show that CovNorm with a global PCA achieves higher recognition rates with $7\times$ less parameters.

\section{MDL by covariance normalization}

In this section, we introduce the CovNorm procedure for MDL with
deep networks.

\subsection{Multi-domain learning}

Figure~\ref{fig:intro} a) motivates the use of
$\bf{A}$ layers in MDL. The figure depicts two
fixed weight layers, ${\bf F}_1$ and ${\bf F}_2$,
and a non-linear layer ${\bf \phi}(.)$ in between.
Since the fixed layers are pre-trained on a {\it source\/} dataset ${\cal S}$, 
typically ImageNet, all weights are optimized for the source statistics. 
For standard losses, such as cross entropy, this is a maximum likelihood (ML)
procedure that matches ${\bf F}_1$ and ${\bf F}_2$ to the
statistics of activations
${\bf x}, {\bf y}$ and ${\bf u}$ in $\cal S$. However,
when the CNN is used on a different {\it target\/}
domain, the statistics of these variables change and
${\bf F}_1, {\bf F}_2$ are no longer an ML solution. Hence, the network is
sub-optimal and must be finetunned on a target dataset ${\cal T}$.
This is denoted full network finetuning and converts the network into an ML 
solution for ${\cal T}$, with the
%It consists of running a few more iterations of the ML 
%procedure, using the network of {\bf a)} as initialization. This has the 
outcome of Figure~\ref{fig:intro} b). In the target domain,  the 
intermediate random variables become ${\bf x}'$, ${\bf y}',$ and ${\bf u}'$
and the weights are changed accordingly, into ${\bf F'}_1$ and 
${\bf F'}_2$.

While very effective, this procedure has two drawbacks, which follow from 
updating all weights. First, it can be
computationally expensive, since modern CNNs have large weight
matrices. Second, because the weights ${\bf F'}_i$ are not optimal
for ${\cal S}$, i.e. the CNN forgets the source task, there is a need
to store and implement two CNNs to solve both tasks. 
This is expensive in terms of storage and computation and increases
the complexity of managing the network ecology.
A device that solves both tasks must store two CNNs and load them in and 
out of cache when it switches between the tasks. 
These problems are
addressed by the MDL architecture of Figure \ref{fig:BlockD},
which is replicated in greater detail on Figure \ref{fig:intro} c). 
It introduces an  {\it adaptation layer\/} ${\bf A}$ and 
fine-tunes this layer only, leaving ${\bf F}_1$ and ${\bf F}_2$ unchanged. 
In this case, the statistics of the input are still those of ${\bf x}'$, but 
the distributions along the network are now those of 
${\bf z}', {\bf y}'',$ and ${\bf u}''$. Since ${\bf F}_1$ is
fixed, nothing can be done about ${\bf z}'$. However, the fine-tuning of
${\bf A}$ encourages the statistics of ${\bf y}''$ to match those
of ${\bf y}'$, i.e. ${\bf y}'' = {\bf y}'$ and thus ${\bf u}'' = {\bf u}'$. 
Even if $\bf A$ cannot match statistics exactly, the mismatch is
reduced by repeating the procedure in subsequent layers, e.g.
introducing a second ${\bf A}$ layer after ${\bf F}_2$, and 
optimizing adaptation matrices as a whole.

\subsection{Adaptation layer size}

Obviously, MDL has limited interest if $\bf A$ has size similar
to ${\bf F}_1$. In this case, each domain has as many adaptation parameters
as the original network, all networks have twice the size, task switching
is complex, and training complexity is equivalent to full fine tunning of 
the original network. On the  other hand, if $\bf A$ is much smaller than 
${\bf F}_1$, MDL  is computationally light and task-switching much more 
efficient.
% The implementation of networks for $D$ domains requires
% a total of $N+Dn$ parameters, which can still be approximately equal to
% $N$. ]
 In summary, the goal is to introduce an adaptation layer ${\bf A}$ 
as {\it small as possible,\/} but still powerful enough to {\it match the
statistics\/} of ${\bf y}'$ and ${\bf y}''$. A simple solution is to 
make $\bf A$ a batch normalization layer~\cite{ioffe2015batch}. This was 
proposed  in~\cite{bilen2017universal} but, as discussed below, is not effective. 
%The problem is
%that, while it can match the means and variances of individual features,
%batch normalization cannot match the statistical dependences of the
%two distributions. This is not a problem during standard network
%traning, where the subsequent layers of the network are changing and
%can absorb the mismatch. It is, however, a major limitation for a adaptation
%layers, since the subsequent layer ${\bf W}_2$ of the network are fixed.
%Variations of the joint statistics of ${\bf z}'$ can lead to substantial drop
%in overall performance, even if the marginal distributions are matched.
To overcome this problem, \cite{rosenfeld2017incremental} proposed a
linear transformation $\bf A$ and \cite{rebuffi2017learning} 
adopted the residual structure of~\cite{he2016deep}, i.e. an adaptation 
layer ${\bf T} = ({\bf I} + {\bf A})$. To maximize parameter savings, 
${\bf A}$ was implemented with a $1 \times 1$ convolutional layer in both
cases. 
% \cite{rebuffi2017learning} showed that residual adaptation layers 
% with $10\%$ of the original CNN size can outperform full network finetuning.

This can, however, still require a non-trivial number of parameters, 
especially in upper network layers. Let ${\bf F}_1$ convolve
a bank of $d$ filters of size $k \times k \times l$ with $l$ feature 
maps. Then, ${\bf F}_1$ has size $d k^2 l$,
${\bf y}$ is $d$ dimensional, and ${\bf A}$ a $d \times d$ matrix. Since in 
upper network layers $k$ is usually small and $d > l$, $\bf A$ can be
only marginally smaller than ${\bf F}_1$. \cite{rebuffi2018efficient}  exploited
redundancies across tasks to address this problem,
creating a matrix with the $\bf A$ layer parameters of multiple tasks
and computing a low-rank approximation of this matrix with an SVD.
The compression achieved with this approximation is limited,
because the approximation is purely geometric, not taking into account the 
statistics of ${\bf z}'$ and ${\bf y}'$. In this work, we propose a more 
efficient solution, motivated by the interpretation of $\bf A$ as 
converting the statistics of ${\bf z}'$ into those of ${\bf y}'$. It
is assumed that the fine-tuning of ${\bf A}$ produces an output 
variable ${\bf y}''$ whose statistics match those of ${\bf y}'$.
This could leverage adaptation layers
in other layers of the network, but that is not important for the discussion
that follows. The only assumption is that ${\bf y}'' = {\bf y'}$.
The goal is to replace $\bf A$ by a simpler matrix that maps ${\bf z}'$
into ${\bf y}'$. For simplicity, we drop the primes and notation of
Figure~\ref{fig:intro} in what follows, considering the problem of matching
statistics between input $\bf x$ and output $\bf y$ of a matrix $\bf A$.

\subsection{Geometric approximations}
\label{sec:geo_appr}

One possibility is to use a purely geometric
solution~\cite{rebuffi2018efficient}. 
Geometrically, the closest low rank approximation of a matrix $\bf A$ is 
given by the SVD, ${\bf A} = {\bf USV}^T$.
More precisely, the minimum Frobenius norm approximation
$\tilde{\bf A} = \arg\min_{\{{\bf B}|rank ({\bf B}) = r\}}
||{\bf A} - {\bf B}||^2_F$, where $r < rank({\bf A})$, is
$\tilde{\bf A} = {\bf U}\tilde{\bf S}{\bf V}^T$ where $\tilde{\bf S}$
contains the $r$ largest singular values of $\bf A$. This
can be written as $\tilde{\bf A} = {\bf C}{\bf W}$, where
${\bf C} = {\bf U}\sqrt{\tilde{\bf S}}$ and
${\bf W} = \sqrt{\tilde{\bf S}}{\bf V}^T$. If $A \in \mathbb{R}^{d \times d}$,
these matrices have a total of $2rd$ parameters. 
An even simpler solution is to define ${\bf C} \in
\mathbb{R}^{d \times r}$ and ${\bf W} \in \mathbb{R}^{r \times d}$,
replace ${\bf A}$ by their product in Figure~\ref{fig:intro} c),
and fine-tune the two matrices instead of $\bf A$. We denote
this as the {\it fine-tunned approximation\/} (FTA).
These approaches are limited by their purely geometric nature. 
Note that $d$ is determined by the source model
(output dimension of ${\bf F}_1$) and fixed. On the other hand,
the dimension $r$ should depend on the target dataset $\cal T$.
Intuitively, if $\cal T$ is much smaller than $\cal S$, or if the target
task is much simpler, it should be possible to use a smaller $r$ than
otherwise. There is also no reason to believe that a single $r$, or even a
single ratio $r/d$, is suitable for all network layers. While $r$ could be
found by cross-validation, this becomes expensive when there
are multiple adaptation layers throughout the CNN. We next introduce 
an alternative, data driven, procedure that bypasses these difficulties.

\begin{figure}[t]
  \begin{minipage}{1.0\linewidth}
    \includegraphics[width=.9\linewidth]{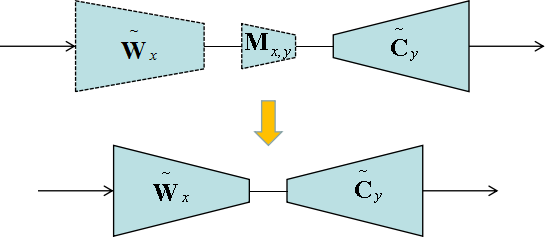}
    \caption{Top: covnorm approximates adaptation
      layer $\bf A$ by a sequence of whitening $\tilde{\bf W}_x$, 
      mini-adaptation ${\bf M}_{x,y}$, and coloring $\tilde{\bf C}_y$
      operations. Bottom: after covnorm, the mini adaptation layer can 
      be absorbed into $\tilde{\bf W}_x$ (shown in the figure) or
      $\tilde{\bf C}_y$.}
    \label{fig:residual_pca_MAL}
  \end{minipage}
\end{figure}

\subsection{Covariance matching}

Assume that, as illustrated in Figure~\ref{fig:covnorm},
$\bf x$ and $\bf y$ are Gaussian random variables of
means $\mu_x, \mu_y$ and covariances ${\bf \Sigma}_x, {\bf \Sigma}_y$,
respectively, related by ${\bf y} = {\bf Ax}$. Let the covariances have
eigendecomposition
\begin{equation}
  {\bf \Sigma}_x = {\bf P}_x {\bf E}_{x}{\bf P}^T_x 
  \quad \quad
  {\bf \Sigma}_y = {\bf P}_y {\bf E}_{y}{\bf P}^T_y
  \label{eq:singvs}
\end{equation}
where ${\bf P}_x, {\bf P}_y$ contain eigenvectors as columns and
${\bf E}_x,{\bf E}_y$ are diagonal eigenvalue matrices. We refer to
the triplet ${\cal P}_x = ({\bf P}_x, {\bf E}_x, \mu_x)$ as the
PCA of $\bf x$. 
Then, it is well known that the statistics of $\bf x$ and $\bf y$ are
related by
\begin{equation}
{\bf \mu}_y = {\bf A}{\bf \mu}_x 
\quad \quad
{\bf \Sigma}_y = {\bf A} {\bf \Sigma}_x {\bf A}^T
\label{eq:Sy}
\end{equation}
and, combining~(\ref{eq:singvs}) and~(\ref{eq:Sy}),
${\bf P}_y {\bf E}_{y}{\bf P}^T_y = {\bf AP}_x
{\bf E}_{x}{\bf P}^T_x{\bf A}^T$. This holds when
${\bf P}_y \sqrt{\bf E}_{y} ={\bf AP}_x \sqrt{\bf E}_{x}$ or, equivalently,
\begin{eqnarray}
  {\bf A} &=& {\bf P}_y\sqrt{{\bf E}_y} \sqrt{{\bf E}^{-1}_x}{\bf P}^T_x.
              \label{eq:A} \\
  &=& {\bf C}_y {\bf W}_x \label{eq:Adecomp}
\end{eqnarray}
where ${\bf W}_x = \sqrt{{\bf E}^{-1}_x}{\bf P}^T_x$%    \label{eq:W}
 is the ``whitening matrix'' of $\bf x$ and
${\bf C}_y = {\bf P}_y\sqrt{{\bf E}_y}$% \label{eq:C}
 the ``coloring matrix'' of $\bf y$. It follows that (\ref{eq:Sy}) holds
if ${\bf y} = {\bf Ax}$ is implemented with a sequence of two operations.
First, ${\bf x}$ is mapped into a variable $\bf w$ of zero mean and
identity covariance, by defining
\begin{eqnarray}
  {\bf w} &=& {\bf W}_x({\bf x} - {\bf \mu}_x). \label{eq:whit}
\end{eqnarray}
Second, $\bf w$ is mapped into $\bf y$ with
\begin{eqnarray}
  {\bf y} &=& {\bf C}_y{\bf w} + {\bf \mu}_y. \label{eq:color}
\end{eqnarray}
In summary, for Gaussian ${\bf x}$, the effect of ${\bf A}$ is
simply the combination of a whitening of ${\bf x}$ followed by a
colorization with the statistics of ${\bf y}$.

\subsection{Covariance normalization}
\label{sec:cov_norm}

The interpretation of the adaptation layer as a recoloring operation
(whitening + coloring) sheds light on the number of parameters
effectively needed for the adaptation, since the PCAs ${\cal P}_x,{\cal P}_y$
capture the {\it effective\/} dimensions of $\bf x$ and $\bf y$.
Let $k_x$ ($k_y$) be the number of eigenvalues significantly larger
than zero in ${\bf E}_x$ (${\bf E}_y$). Then, the whitening and
coloring matrices can be approximated by
\begin{eqnarray}
  \tilde{{\bf W}}_x = \sqrt{\tilde{{\bf E}}^{-1}_x} \tilde{{\bf P}}^T_x \quad \quad
  \tilde{{\bf C}}_y = \tilde{{\bf P}}_y \sqrt{\tilde{{\bf E}}_y}
  \label{eq:tildes}
\end{eqnarray}
where $\tilde {{\bf E}}_x \in \mathbb{R}^{k_x \times k_x}$
($\tilde {{\bf E}}_y \in \mathbb{R}^{k_y \times k_y}$) contains the
non-zero eigenvalues of ${\bf \Sigma}_x$ (${\bf \Sigma}_y$),
and $\tilde{{\bf P}}_x \in \mathbb{R}^{d \times k_x}$
($\tilde{{\bf P}}_y \in \mathbb{R}^{d \times k_y}$)
the corresponding eigenvectors. Hence, ${\bf A}$ is well
approximated by a pair of matrices ($\tilde{{\bf W}}_x$,
$\tilde{{\bf C}}_y$) totaling $d(k_x + k_y)$ parameters.

On the other hand, the PCAs are only defined up to a permutation,
which assigns an ordering to eigenvalues/eigenvectors.
When the input and output PCAs are computed
independently, the principal components may not be aligned.
This can be fixed by introducing a permutation matrix between
${\bf C}_y$ and ${\bf W}_x$ in~(\ref{eq:Adecomp}).
The assumption that all distributions are Gaussian also only holds
approximately in real networks. To account for all this, we augment the
recoloring operation with a mini-adaptation layer
${\bf M}_{x,y}$  of size $k_x \times k_y$. This leads to the
covariance normalization (CovNorm) transform
\begin{eqnarray}
  \tilde{\bf y} = \tilde{\bf C}_y {\bf M}_{x,y} \tilde{\bf W}_x ({\bf x}
  - \mu_x) + \mu_y,
  \label{eq:adapt}
\end{eqnarray}
where ${\bf M}_{x,y}$ is learned by fine-tuning on the target
dataset $\cal T$. Beyond improving recognition performance, this has the
advantage of further parameters savings. The direct
implementation of (\ref{eq:adapt}) increases the parameter count to
$d(k_x + k_y) + k_xk_y$. However, after fine-tuning, ${\bf M}_{x,y}$
can be absorbed into one of the two other matrices , as shown in Figure 
\ref{fig:residual_pca_MAL}. When $k_x > k_y$, ${\bf M}_{x,y}\tilde{\bf W}_x$ 
has dimension $k_y \times d$ and replacing the two matrices by their product 
reduces the total parameter count to $2 d k_y$. In this case, we say 
that ${\bf M}_{x,y}$ is absorbed into $\tilde{\bf W}_x$. Conversely, 
if $k_x < k_y$, ${\bf M}_{x,y}$ can be absorbed into $\tilde{\bf C}_y$. 
Hence, the total parameter count is $2 d \min(k_x, k_y)$. CovNorm is 
summarized in Algorithm~\ref{algo:covnorm}.

\begin{algorithm}[t]
	\KwData{source $\cal S$ and target $\cal T$}
	Insert an adaptation layer ${\bf A}$ on a CNN trained on $\cal S$ 
        and fine-tune $\bf A$ on ${\cal T}$.
	
	Store the layer input and output PCAs ${\cal P}_x$, ${\cal P}_y$,
	select the $k_x, k_y$ non-zero eigenvalues and corresponding
        eigenvectors from each PCA, and compute $\tilde{\bf C}_y,
        \tilde{\bf W}_x$ with (\ref{eq:tildes}).
        
	add mini-adaptation layer ${\bf M}_{x,y}$ and replace $\bf A$
	by (\ref{eq:adapt}). Note that, as usual, the constant 
	$\tilde{\bf C}_y {\bf M}_{x,y} \tilde{\bf W}_x \mu_x + \mu_y$ can be
	implemented with a vector of biases.
	
	fine-tune ${\bf M}_{x,y}$ with $\tilde{\bf W}_x$ and $\tilde{\bf C}_y$ on $\cal T$ and absorb ${\bf M}_{x,y}$ into
	the larger of $\tilde{\bf W}_x$ and $\tilde{\bf C}_y$.
	\caption{Covariance Normalization }
	\label{algo:covnorm}
\end{algorithm}

\subsection{The importance of covariance normalization}
\label{sec:impor_cov_norm}

The benefits of covariance matching can be seen by comparison
to previously proposed MDL methods. Assume, first, that $\bf x$
and $\bf y$ consist of {\it independent\/} features. In this case,
${\bf P}_x, {\bf P}_y$ are identity matrices and
(\ref{eq:whit})-(\ref{eq:color}) reduce to
\begin{eqnarray}
  y_i = \sqrt{e_{y,i}} \frac{x_i - \mu_{x,i}}{\sqrt{e_{x,i}}} + \mu_{y,i},
  \label{eq:bn}
\end{eqnarray}
which is the batch normalization equation. Hence, CovNorm is a generalized 
form of the latter. There are, however, important
differences. First, there is no batch. The normalizing distribution
$\bf x$ is now the distribution of the feature responses of 
layer ${\bf F}_1$ on the target dataset $\cal T$. Second, the goal is
not to facilitate the learning of ${\bf F}_2$, but produce
a feature vector $\bf y$ with statistics matched to ${\bf F}_2$.
This turns out to make
a significant difference. Since, in regular batch normalization,
${\bf F}_2$ is allowed to change, it can absorb any initial mismatch
with the independence assumption. This is not
the case for MDL, where ${\bf F}_2$ is {\it fixed.\/}
Hence,~(\ref{eq:bn}) usually fails, significantly underperforming
(\ref{eq:whit})-(\ref{eq:color}).

Next, consider the geometric solution.
Since CovNorm reduces to the product of two tall
matrices, e.g. ${\bf K} = \tilde{\bf C}_y {\bf M}_{x,y}$
and ${\bf L} = \tilde{\bf W}_x$ of size $d \times k_x$, it should be
possible to replace it with the fine-tuned approximation based on two 
matrices of this size. Here, there are two difficulties. First, $k_x$ is 
not known in the absence of the PCA decompositions. Second, in our 
experience, even when $k_x$ is set to the value used by PCA, the
fine-tuned approximation does not work. As shown in the
experimental section, when the matrices are initialized with
Gaussian weights, performance can decrease significantly.
%PCA
%initialization appears to be critical to achieve performance comparable to
%that of $\bf A$.
This is an interesting observation
because $\bf A$ is itself initialized with Gaussian weights.
It appears that a good initialization is more critical
for the low-rank matrices.

Finally, CovNorm can be compared to the SVD, ${\bf A} = {\bf USV}^T$.
%Using this form in (\ref{eq:Sy}) and
%using (\ref{eq:singvs}) leads to
%\begin{eqnarray}
%  {\bf \Sigma}_y
%  &=& {\bf USV}^T {\bf \Sigma}_x {\bf VS}^T{\bf U}^T \\
%  &=& {\bf USV}^T {\bf P}_x {\bf E}_x {\bf P}_x^T {\bf VS}^T{\bf U}^T \\
% {\bf P}_y\sqrt{\bf E_y} &=& {\bf USV}^T {\bf P}_x \sqrt{\bf E}_x.
                              %  \end{eqnarray}
From~(\ref{eq:A}), this holds whenever ${\bf V} = {\bf P}_x$,
${\bf S} = \sqrt{{\bf E}_y}\sqrt{{\bf E}^{-1}_x}$ and
${\bf U} = {\bf P}_y$. The problem is that the singular value matrix $\bf S$
conflates the variances of the input and output PCAs. The fact that $s_i =
e_{y,i}/e_{x,i}$ has two important consequences. First, it is impossible
to recover the dimensions $k_x$ and $k_y$ by inspection of the
singular values. Second, the low-rank criteria of selecting the largest
singular values is {\it not\/} equivalent to CovNorm.
For example, the principal components of $\bf x$
with largest eigenvalues $e_{x,i}$ have the smallest singular values
$s_i$. Hence, it is impossible to tell if singular vectors ${\bf v}_i$ of
small singular values are the most important (PCA components of
large variance for $\bf x$) or the least important (noise).
Conversely, the largest singular values can simply signal the
least important input dimensions. CovNorm eliminates
this problem by explicitly selecting the important input and
output dimensions.

\subsection{Joint training}

\cite{rebuffi2018efficient} considered a variant of MDL where the different tasks of
Figure~\ref{fig:BlockD} are all optimized simultaneously. This is the
same as assuming that a joint dataset ${\cal T} = \cup_i {\cal T}_i$ is
available. For CovNorm, the only difference with respect to the single
dataset setting
is that the PCAs ${\cal P}_x, {\cal P}_y$ are now those of the joint data
${\cal T}$. These can be derived from the PCAs
${\cal P}_{x,i}, {\cal P}_{y,i}$ of the individual target datasets ${\cal T}_i$
with
\begin{eqnarray}
  {\bf \mu}_{\cal T} &=& \frac{1}{N}\sum_i N_i \mu_i \nonumber \\
  {\bf \Sigma}_{\cal T} &=& \sum_i \frac{N_i}{N}
  ({\bf P}_i {\bf E}_i {{\bf P}_i}^T + \mu_i {\mu_i}^T))
  - {\bf \mu}_{\cal T}{\bf \mu}_{\cal T}^T
  \label{eq:cov_accum}
\end{eqnarray}
where $N_i$ is the cardinality of ${\cal T}_i$.
% and covariance
% \begin{eqnarray}
%   {\bf \Sigma}_{\cal T}
%   &=& E[x_{{\cal T}}^2] - {\bf \mu}_{\cal T}{\bf \mu}_{\cal T}^T \\
%   &=&  \sum_i \frac{N_i}{N} E[x_{{\cal T}_i}^2] - {\bf \mu}_{\cal T}{\bf \mu}_{\cal T}^T \\
%   &=&  \sum_i \frac{N_i}{N} ({\bf \Sigma}_{{\cal T}_i} + \mu_i {\mu_i}^T) - {\bf \mu}_{\cal T}{\bf \mu}_{\cal T}^T \\
%   &=& \sum_i \frac{N_i}{N} ({\bf P}_i {\bf E}_i {{\bf P}_i}^T + \mu_i {\mu_i}^T)) - {\bf \mu}_{\cal T}{\bf \mu}_{\cal T}^T
%  \label{eq:cov_accum}
%\end{eqnarray}
Hence, CovNorm can be implemented by finetuning ${\bf A}$ to each 
${\cal T}_i$, storing the PCAs ${\cal P}_{x,i}, {\cal P}_{y,i}$,
using (\ref{eq:cov_accum}) to
reconstruct the covariance of $\cal T$, and computing the
global PCA. When tasks are available sequentially,
this can be done recursively, combining the PCA of 
all previous data with the PCA of the new data. In summary, CovNorm
can be extended to any number of tasks, with constant
storage requirements (a single PCA), and no loss of optimality.
This makes it possible to define two CovNorm {\it modes\/}.
\begin{itemize}
\item {\it independent:} $\bf A$ layers of network $i$ are adapted to target
  dataset ${\cal T}_i$. A PCA is computed for ${\cal T}_i$ and the
  mini-adaptation fine-tuned to ${\cal T}_i$. This
  requires $2d\min(k_x,k_y)$ task specific parameters (per layer) per dataset. 
\item {\it joint:} a global PCA is learned from
  $\cal T$ and $\tilde{\bf C}_y, \tilde{\bf W}_x$ shared
  across tasks. Only a mini-adaptation layer is fine-tuned per ${\cal T}_i$.
  This requires $\min(k_x,k_y)$ task-specific parameters (per layer) per
  dataset. All ${\cal T}_i$ must be available simultaneously.
% \item {\it learning without forgetting:} same as universal model, but 
%   tasks are introduced sequentially. Hence, although using a ``global PCA'', 
%   earlier models use a PCA that accounts for less target datasets and is more 
%   representative of their own statistics. Note that when a new target 
%   dataset is added, the new model accounts for all previous tasks, and can 
%   be used to solve all of them. Hence, there are no  additional parameters 
%   per dataset.   All ${\cal T}_i$ must be kept and available simultaneously.
% \item {\it learning with almost no forgetting:} Same as learning
%   without forgetting, but the model includes a mini-adaptation layer
%   fine-tuned on each target dataset. This requires $k_xk_y$ parameters.
%   Note that, once a target dataset is added and the global PCA updated,
%   the mini-adaptation matrices of all previous models have to be fine-tuned.
%   Hence, All ${\cal T}_i$ must be kept and available 
%   simultaneously.
\end{itemize}
The independent model is needed if, for example, the devices of
Figure~\ref{fig:BlockD} are produced by different manufacturers. 

\begin{figure*}[t]\RawFloats
  \begin{minipage}{0.49\linewidth}
    \includegraphics[width=.49\linewidth]{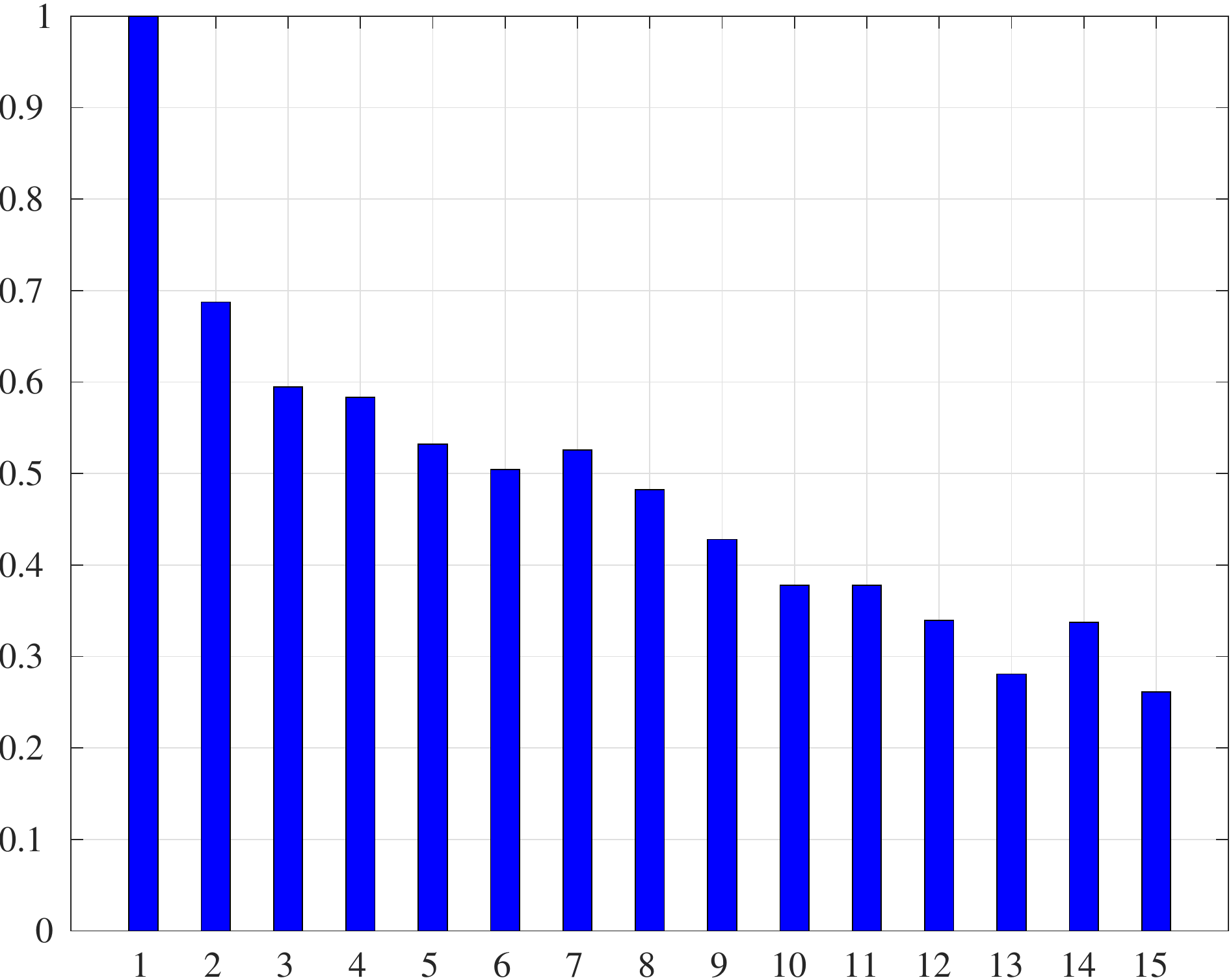} 
    \includegraphics[width=.49\linewidth]{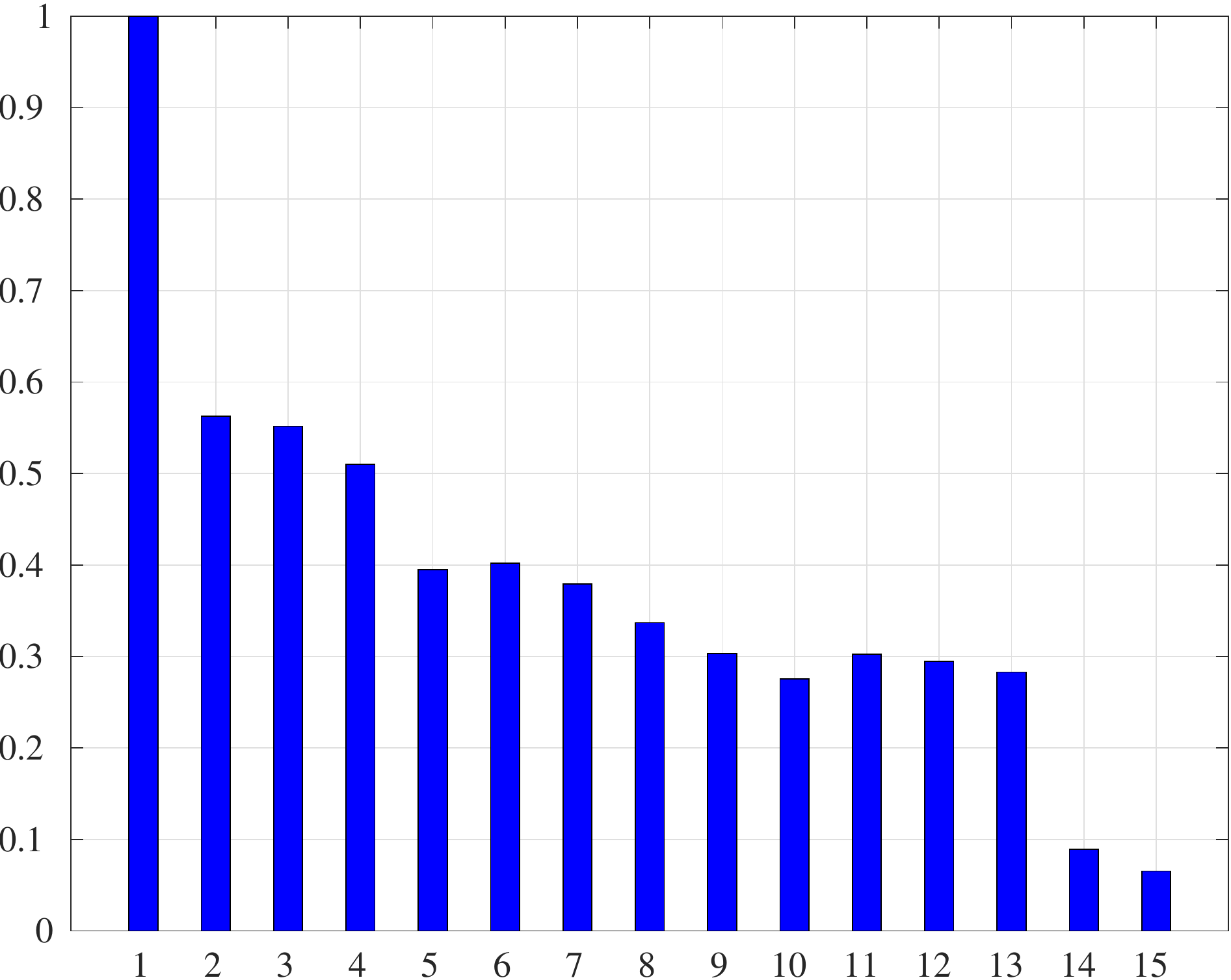}
    \caption{Ratio of effective dimensions ($\eta$) for different
      network layers. Left: MITIndoor. Right: CIFAR100.}
    \label{fig:ratio}
  \end{minipage} \,\,
  \begin{minipage}{.49\linewidth}
    \includegraphics[width=.49\linewidth]{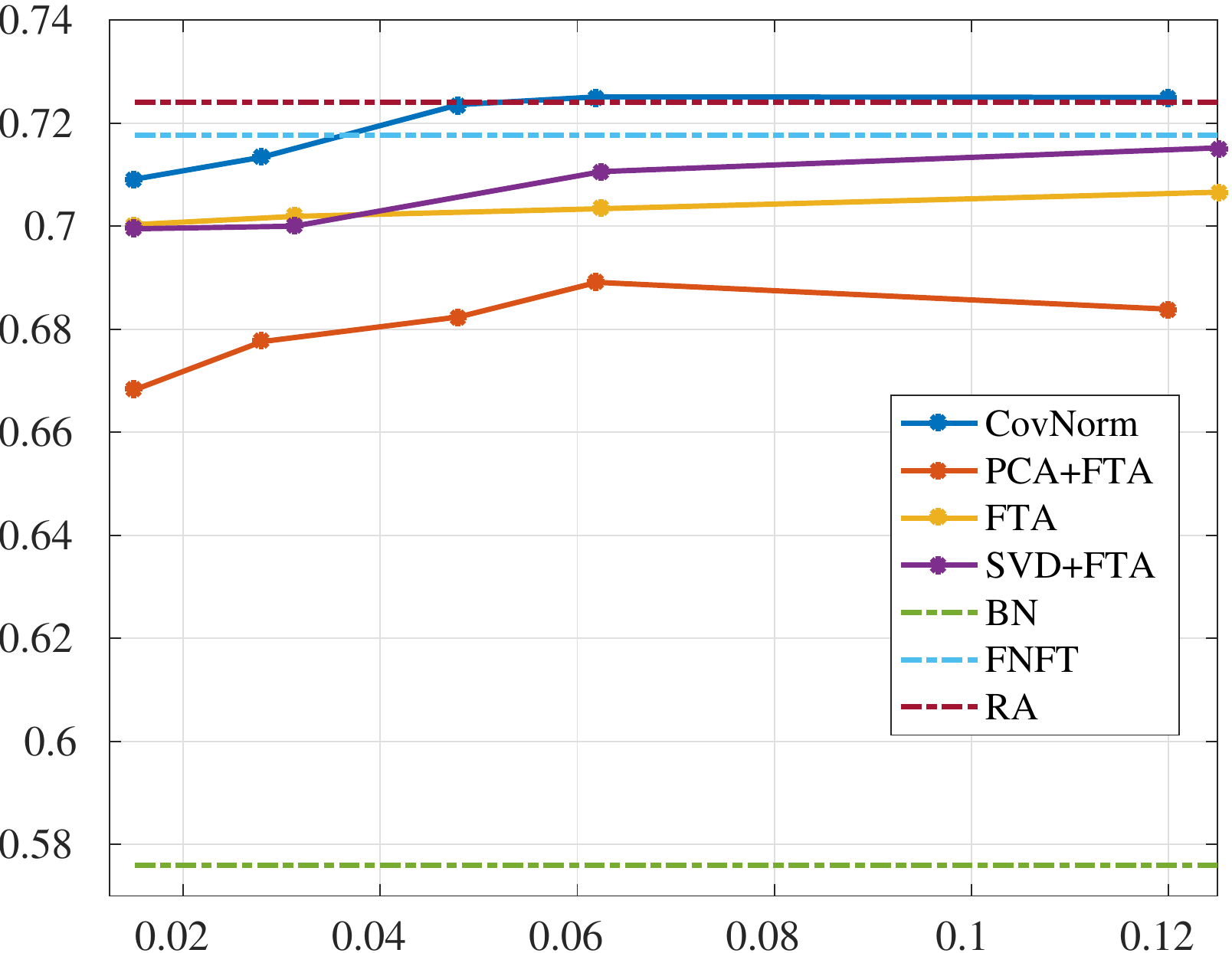} 
    \includegraphics[width=.49\linewidth]{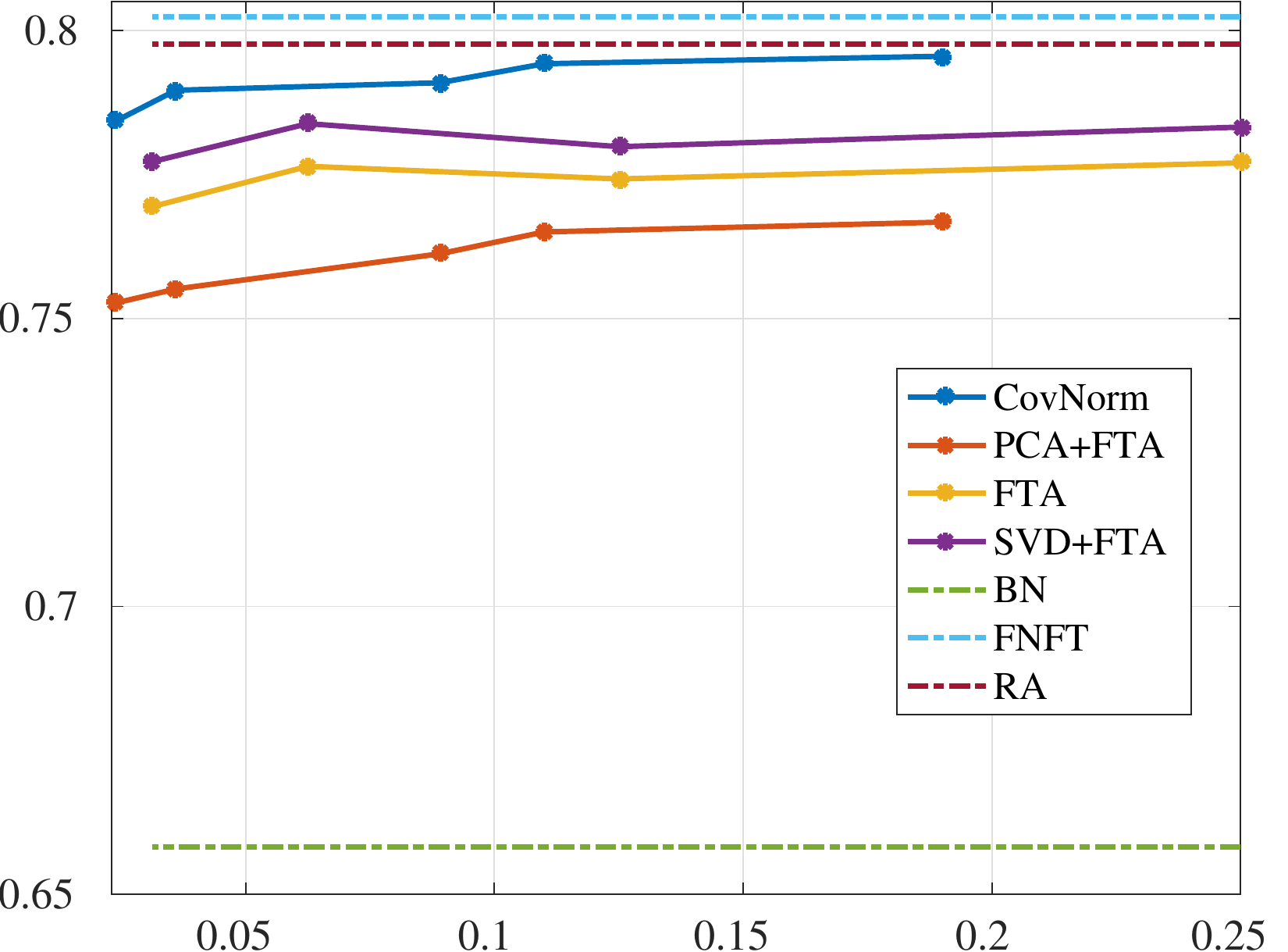}
    \caption{accuracy vs. \% of parameters used for 
      adaptation. Left: MITIndoor. Right: CIFAR100.}
    \label{fig:acc}
  \end{minipage}
\end{figure*}

\section{Experiments}
In this section, we present results for both the independent and joint
CovNorm modes.

{\bf Dataset:} \cite{rebuffi2017learning} proposed the decathlon dataset for 
evaluation of MDL. However, this is a collection of relatively small datasets.
While sufficient to train small networks,
we found it hard to use with larger CNNs. Instead,
we used a collection of seven popular vision datasets.
{\bf SUN 397}~\cite{xiao2010sun} 
contains $397$ classes of scene images and more than a million images. 
{\bf MITIndoor} \cite{valenti2007indoor} is an indoor scene dataset
with $67$ classes and $80$ samples per class.  
{\bf FGVC-Aircraft Benchmark}~\cite{maji2013fine} is a fine-grained 
classification dataset of $10,000$ images of $100$ types of airplanes. 
{\bf Flowers102} \cite{nilsback2008automated} is a fine-grained dataset
with $102$ flower categories and $40$ to $258$ images per
class. {\bf CIFAR100}~\cite{krizhevsky2009learning} contains $60,000$ tiny 
images, from $100$ classes. 
%All the samples have the same size of $32\times 32$ and there are $600$ samples for each class. 
{\bf Caltech256}~\cite{griffin2007caltech} contains $30,607$ images
of $256$ object categories, with at least $80$ samples per class. 
{\bf SVHN} \cite{netzer2011reading} is a digit recognition dataset with
$10$ classes and more than $70,000$ samples. In all cases, images are resized 
to $224\times 224$ and the training and testing splits defined by the
dataset are used, if available. Otherwise, $75\%$ is used for training 
and $25\%$ for testing.

{\bf Implementation:} In all experiments, fixed $\bf F$ layers
were extracted from a source {\bf VGG16} \cite{simonyan2014very} model
trained on ImageNet. This has convolution layers of dimensions
ranging from $64$ to $4096$. In a set of preliminary
experiments, we compared the MDL performance of the architecture of
~\ref{fig:BlockD} with these $\bf F$ layers and adaptation
layers implemented with 1) a  convolutional layer $\bf A$ of kernel
size $1\times 1$ ~\cite{rosenfeld2017incremental}, 2) the residual adapters
${\bf T}={\bf B_2}({\bf I} + {\bf AB_1})$ of \cite{rebuffi2017learning},
where ${\bf B_1}$ and ${\bf B_2}$ are batch normalization layers and ${\bf A}$
as in 1), and 3) the parallel adapters of~\cite{rebuffi2018efficient}. Since residual
adapters produced the best results, we adopted this structure in all our
experiments. However, CovNorm can be used with any of the other
structures, or any other matrix $\bf A$. Note that
${\bf B}_1$ could be absorbed into ${\bf A}$ after fine-tuning but we have
not done so, for consistency with \cite{rebuffi2017learning}. 
% the overall adaptive block that can convert the original convolutional layer ${\bf M}$ to ${\bf M'}$ for the target dataset. As both ${\bf B_1}$ and ${\bf B_2}$ are diagonal matrices and ${\bf B_1}$ can always be absorbed by ${\bf TB_1}$, we mainly discuss ${\bf T}$ in the following part.
%During the training stage, ${\bf B_1}$, ${\bf B_2}$ and ${\bf T}$ are 
%changed. 

In all experiments, fine-tuning used initial learning rate of $0.001$,
reduced by $10$ when the loss stops decreasing. After fine-tuning
the residual layer, features were extracted at the input and
output of ${\bf A}$ and the PCAs ${\cal P}_x, {\cal P}_y$ computed and used in
Algorithm~\ref{algo:covnorm}. Principal components were selected by the
explained variance criterion. Once the eigenvalues $e_i$ were computed
and sorted by decreasing magnitude, i.e. $e_1\geq e_2 \geq \ldots \geq e_d$,
the variance explained by the first $i$ eigenvalues is
$r_i = \frac{\sum_{k=1}^i e_i}{\sum_{k=1}^d e_i}$. 
Given a threshold $t$, the smallest index $i^*$ such that $r_{i^*} > t$ was
determined, and only the $i^*$ first eigenvalues/eigenvectors 
were kept. This set the dimensions $k_x, k_y$ (depending
on whether the procedure was used on ${\cal P}_x$ or ${\cal P}_y$).
Unless otherwise noted, we used $t=0.99$, i.e. $99\%$ of the variance was
retained.

\begin{figure*} \RawFloats
	\begin{minipage}{0.7\linewidth}
		\scriptsize
		\centering
		\captionof{table}{\footnotesize{Classification accuracy and $\%$ of adaptation
				parameters (with respect to VGG size) per target dataset.}}
		%\captionsetup[Table]{position=top}
		%\caption{\footnotesize{Classification accuracy and $\%$ of adaptation
		%		parameters (with respect to VGG size) per target dataset.}}
		\label{tab:comparison_result}
		\begin{tabular}{|c|c|c|c|c|c|c|c|c|c|}
			\hline
			&FGVC&MITIndoor&Flowers&Caltech256&SVHN&SUN397&CIFAR100&average\\
			\hline
			FNFT &85.73\%&71.77\%&95.67\%&83.73\%&96.41\%&57.29\%&80.45\%&81.58\%\\
			& \multicolumn{7}{c|}{100\%} & 100\%\\
			\hline\hline
			\multicolumn{9}{|c|}{Independent learning} \\
			\hline\hline 
			BN \cite{bilen2017universal}&43.6\%&57.6\%&83.07\%&73.66\%&91.1\%&47.04\%&64.8\%&65.83\%\\
			& \multicolumn{7}{c|}{0\%} & 0\%\\
			\hline
			LwF\cite{li2017learning}&66.25\%&{\bf 73.43\%}&89.12\%&80.02\%&44.13\%&52.85\%&72.94\%&68.39\%\\
			& \multicolumn{7}{c|}{0\%} & 0\%\\
			\hline
			RA \cite{rebuffi2017learning}&88.92\%&72.4\%&96.43\%&84.17\%&96.13\%& 57.38\%&{\bf 79.55\%}&82.16\%\\
			& \multicolumn{7}{c|}{10\%} & 10\%\\
			\hline
			SVD+FTA &89.07\%&71.66\%&95.67\%&84.46\%&96.04\%&57.12\%&78.28\%&81.75\%\\
			& \multicolumn{7}{c|}{5\%} & 5\%\\
			\hline
			FTA &87.31\%&70.26\%&95.43\%&83.82\%&95.96\%&56.43\%&78.23\%&81.06\%\\
			& \multicolumn{7}{c|}{5\%} & 5\%\\
			\hline
			CovNorm &{\bf 88.98\%}&72.51\%&{\bf 96.76\%}&{\bf 84.75\%}&{\bf 96.23\%}&{\bf 57.97\%}& 79.42\%&{\bf 82.37\%}\\
			&0.34\%&0.62\%&0.35\%&0.46\%&0.13\%&0.71\%&1.1\% & 0.53\%\\
			\hline\hline
			\multicolumn{9}{|c|}{Joint learning} \\
			\hline\hline
			SVD \cite{rebuffi2018efficient}&88.98\%&71.7\%&96.37\%&83.63\%&96\%&56.58\%&78.26\%&81.65\%\\
			& \multicolumn{7}{c|}{5\%} & 5\%\\
			\hline
			CovNorm&{\bf 88.99\%}&{\bf 73.0\%}& {\bf 96.69\%} &{\bf 84.77\%}&{\bf 96.22\%}&\bf 58.2 &\bf 79.22\%&{\bf 82.44\%}\\
			& \multicolumn{7}{c|}{0.51\%} & 0.51\%\\
			\hline
		\end{tabular}
	\end{minipage}
	\begin{minipage}{.29\linewidth}
		\includegraphics[width=\linewidth,height=1.2in]{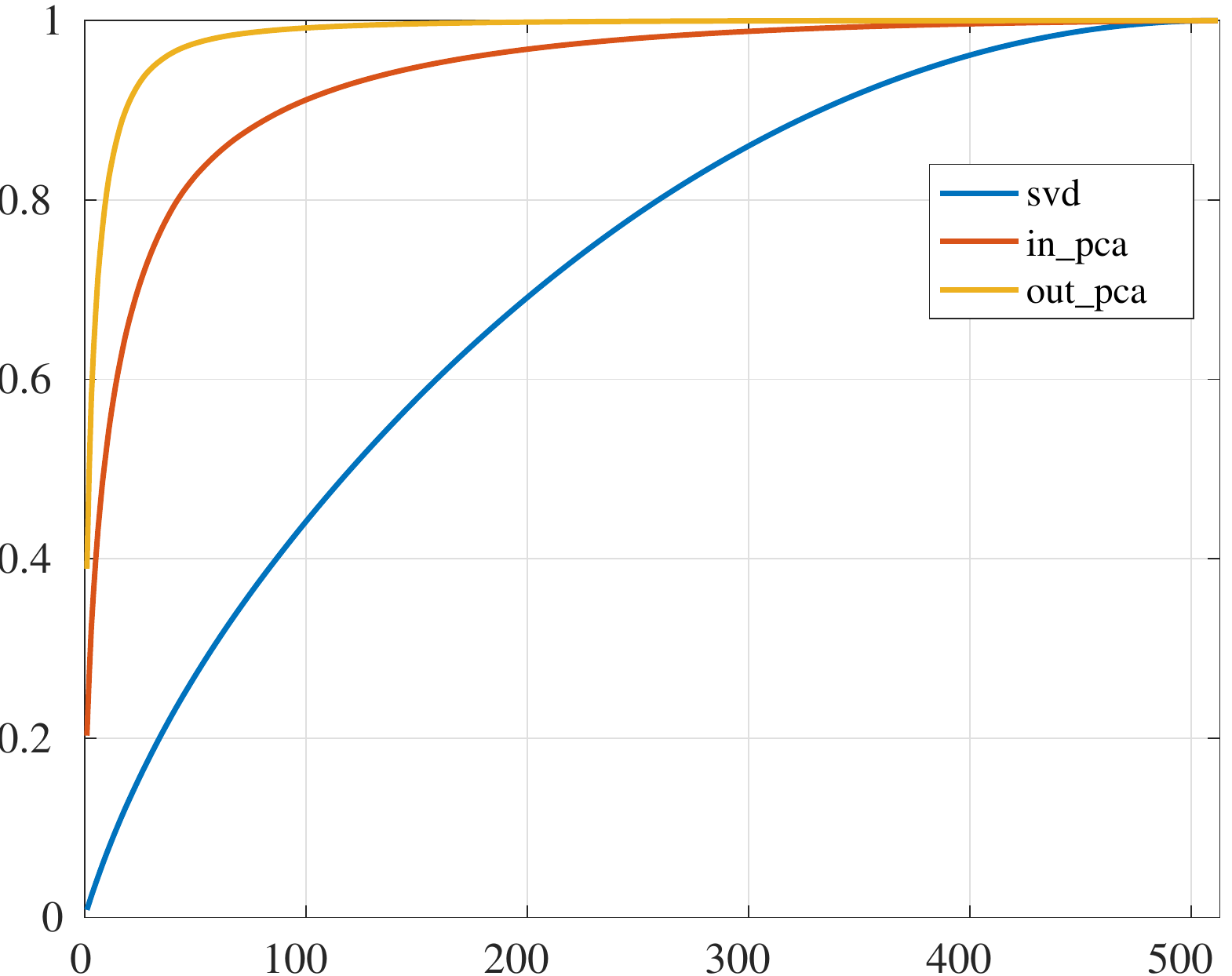} 
		\includegraphics[width=\linewidth,height=1.2in]{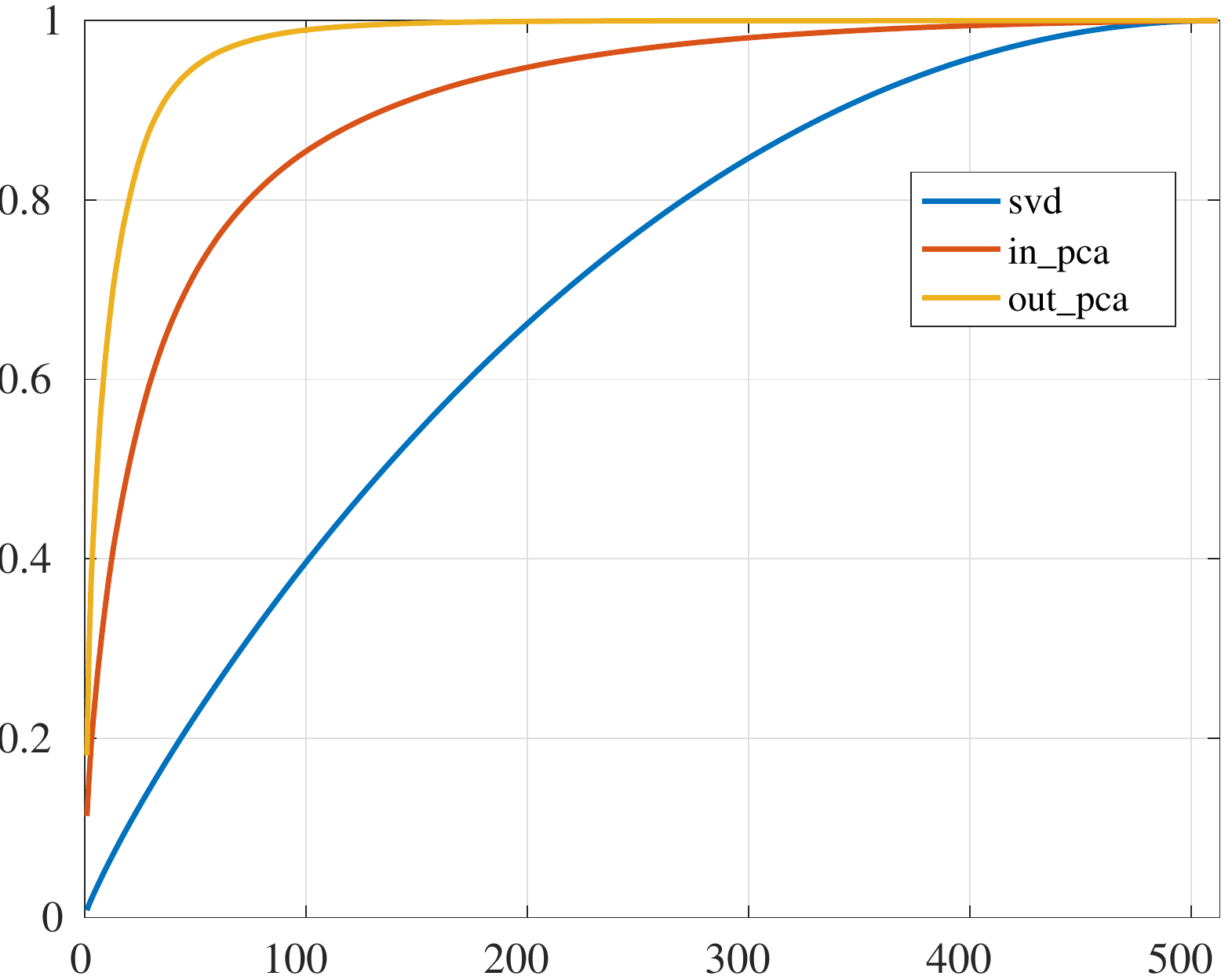}
		\caption{Variance explained by eigenvalues of a layer input and output,
			and similar plot for singular values. Left: MITIndoor. Right: CIFAR100.}
		\label{fig:pca_svd_value}
	\end{minipage}
\end{figure*} 
\begin{table*}
	\centering		
	\scriptsize	
	\begin{tabular}{|c|c|c|c|c|c|c|c|c|c|c|c|c|c|}
		\hline
		&ImNet&Airc&C100&DPed&DTD&GTSR&Flwr&OGlt&SVHN&UCF&avg acc&S&\#par\\
		\hline
		RA \cite{rebuffi2017learning} &59.67\%&61.87\%&81.20\%&93.88\%&57.13\%& 97.57\%&81.67\%&89.62\%&96.13\%&{\bf{50.12\%}}&76.89\%&2621&2\\
		\hline
		DAN \cite{rosenfeld2018incremental} &57.74\%&64.12\%&80.07\%&91.3\%&56.54\%& 98.46\%&{\bf{86.05\%}}&{\bf{89.67\%}}&96.77\%&49.38\%&77.01\%&2851&2.17\\
		\hline
		Piggyback \cite{mallya2018piggyback}&57.69\%&65.29\%&79.87\%&96.99\%&57.45\%& 97.27\%&79.09\%&87.63\%&{\bf{97.24\%}}&47.48\%&76.6\%&2838&1.28\\
		\hline
		CovNorm &{\bf{60.37\%}}&{\bf{69.37\%}}&{\bf{81.34\%}}&{\bf{98.75\%}}&{\bf{59.95\%}}&{\bf{99.14\%}}&83.44\%&87.69\%&96.55\%&48.92\%&{\bf{78.55\%}}&{\bf{3713}}&1.25\\
		\hline
	\end{tabular}			
	\caption{Visual Decathlon results}	
	\label{tab:decathlon}
\end{table*}
{\bf Benfits of CovNorm:} We start with some independent MDL experiments
that provide insight on the benefits of CovNorm over previous MDL procedures.
While we only report results for MITIndoor and CIFAR100,
they are typical of all target datasets. Figure~\ref{fig:ratio} shows the
ratio $\eta = k_y/k_x$ of effective output to input dimensions, as a
function of adaptation layer. It shows that the input of ${\bf A}$ typically
contains more information than the output. Note that $\eta$ is rarely one,
is almost always less than $0.6$, frequently smaller than $0.3$, and
smallest for the top network layers. 
% The input feature is just the output of ${\bf W}_1$ in Figure ~\ref{fig:intro}. As ${\bf W}_1$ is from the original network which is trained with ImageNet and fixed during fine-tuning. Thus the output of it includes much more information. And the adaptive layer $\bf A$ here just extracts the useful information for the target dataset. Based on this observation, we can try to replace $\bf A$ with a much smaller layer which only contain the useful information for the target dataset.

We next compared CovNorm to batch normalization 
(BN)~\cite{bilen2017universal}, and
geometric approximations based on the fine-tunned approximation (FTA) of
Section~\ref{sec:geo_appr}. We also tested a mix of the
geometric approaches (SVD+FTA), where $\bf A$ was first approximated by
the SVD and the matrices $\bf C$, $\bf W$ finetuned on $\cal T$,
and a mix of PCA and FTA (PCA+FTA), where the mini-adaptation
layer ${\bf M}_{x,y}$ of CovNorm was removed and
$\tilde{\bf C}_y, \tilde{\bf W}_x$ fine-tuned on $\cal T$, to minimize
the PCA alignment problem. All geometric approximations
were implemented with low-rank parameter
values $r = d/2^i$, where $d$ is the dimension of $\bf x$ or $\bf y$ and
$i \in \{2, \ldots, 6\}$. For CovNorm, the explained variance threshold was
varied in $[0.8,0.995]$. 
Figure~\ref{fig:acc} shows recognition accuracies
vs. the \% of parameters. Here, $100\%$  parameters corresponds
the adaptation layers of~\cite{rebuffi2017learning}: a network with residual
adapters whose matrix $\bf A$ is fine-tunned on $\cal T$.
This is denoted RA and shown as an upper-bound. A second
upper-bound is shown for full network fine tuning (FNFT). This requires
$10\times$ more parameters than RA. BN, which requires close to zero
parameters, is shown as a lower bound. 

Several observations are possible. First, all geometric approximations 
underperform CovNorm. For comparable sizes, the
accuracy drop of the best geometric method (SVD+FTA)
is as large as $2\%$. This is partly due to the use of a constant low
rank $r$ throughout the network. This cannot match the effective,
data-dependent, dimensions, which vary across layers (see
Figure~\ref{fig:ratio}). CovNorm eliminates this problem.
We experimented with heuristics for choosing variable ranks but,
as discussed below (Figure~\ref{fig:pca_svd_value}),
could not achieve good performance.
%likely due to the ambiguity of singular
%values with regards to effective dimensions discussed in
%Section~\ref{sec:impor_cov_norm}.
Among the geometric approaches, SVD+FTA outperforms FTA, which has
performance drops in most of datasets.
It is interesting that, while $\bf A$ is fine-tuned with random initialization,
the process is not effective for the low-rank matrices of FTA.
In several datasets, FTA could not match SVD+FTA.

Even more surprising were the weaker results obtained when the random
initialization was replaced by the two PCAs (PCA+FTA). Note the large
difference between PCA+FTA and CovNorm
(up to $4\%$), which differ by the
mini-adaptation layer ${\bf M}_{x,y}$.  This is explained by the
alignment problem of Section~\ref{sec:cov_norm}. Interestingly,
while mini-adaptation layers are critical to overcome this problem,
they are as easy to fine-tune as ${\bf A}$. In
fact, the addition of these layers (CovNorm) often outperformed the 
full matrix $\bf A$ (RA). In some datasets, like MITIndoor, with $4.8\%$ of the parameters,
CovNorm matched the performance of RA, Finally, as previously reported by \cite{rebuffi2017learning}, FNFT
frequently underperformed RA. This is likely due to overfitting.

{\bf CovNorm vs SVD:} 
Figure \ref{fig:pca_svd_value} provides empirical evidence for the
vastly different quality of the approximations produced by CovNorm and 
the SVD. The figure shows a plot of the variance explained by 
the eigenvalues of the input and output distributions of an adaptation 
layer $\bf A$ and the corresponding plot for its singular values. 
Note how the PCA energy is packed into a much smaller number of coefficients
than the singular value energy. This happens because 
PCA only accounts for the subspaces populated by data,
restricting the low-rank approximation to these subspaces.
Conversely, the geometric approximation must approximate the
matrix behavior even outside of these subspaces. Note that the SVD is
not only less efficient in identifying the important dimensions, but
also makes it difficult to determine how many singular values to keep.
This prevents the use of a layer-dependent number of singular
values.

{\bf Comparison to previous methods:}
Table~\ref{tab:comparison_result} summarizes the recognition accuracy and 
$\%$ of adaptation layer parameters vs. VGG
model size ($100\%$ parameters), for various methods. 
All abbreviations are as above. Beyond MDL, we 
compare to learning without forgetting (LwF) \cite{li2017learning} a
lifelong method to learn a model that shares all parameters
% (except the last layer)
among datasets. 
%\cite{li2017learning} only presents results for two
% datasets, we present the results of our implementation on all seven.
The table is split into independent and joint MDL.
For joint learning, CovNorm is implemented
with (\ref{eq:cov_accum}) and compared to the SVD approach
of \cite{rebuffi2018efficient}.
% An adaptation matrix ${\bf A}_i$ is
% computed per target dataset, all matrices stacked into a larger matrix, and
% an SVD computed from the latter. This is divided into two matrices, one
% of which is kept fixed and the other finnetuned per dataset.

Several observations can be made. First, CovNorm adapts
the number of parameters to the task, according to its complexity
and how different it is from the source (ImageNet).
For the simplest datasets, such as the $10$-digit class SVHN, adaptation
can require as few as $0.13\%$
task-specific parameters. Datasets that are more diverse but
ImageNet-like, such as Caltech256, require around $0.46\%$
parameters. Finally, larger adaptation layers are required by datasets that are
either complex or quite different from ImageNet, e.g. scene (MITIndoor,
SUN397) recognition tasks. Even
here, adaptation requires less than $1\%$ parameters.
On average, CovNorm requires $0.53\%$
additional parameters per dataset.

Second, for independent learning, all methods based on residual adapters
significantly outperform BN and LwF. As shown by \cite{rebuffi2017learning},
RA outperforms FNFT. BN is uniformly weak, LwF
performs very well on MITIndoor and Caltech256, but poorly on
most other datasets. Third, CovNorm outperforms even RA, achieving
higher recognition accuracy with $20 \times$ less
parameters. It also outperforms SVD+FTA and FTA by $\approx0.6\%$ and
$\approx 1.3\%$, respectively, while reducing parameter sizes by a factor
of $\approx 10$. On a per-dataset basis, CovNorm outperforms RA
on all datasets other than CIFAR100, and SVD+FTA and FTA on all of
them. In all datasets, the parameter savings are significant.
Fourth, for joint training,
CovNorm is substantially superior to the SVD~\cite{rebuffi2018efficient},
with higher recognition rates in all datasets, gains of up to
$1.62\%$ (SUN397), and close to $10\times$ less parameters.
Finally, comparing independent and joint CovNorm, the latter has slightly
higher recognition for a slightly higher parameter count. Hence,
the two approaches are roughly equivalent.

{\bf Results on Visual Decathlon}
Table \ref{tab:decathlon} presents results on the Decathlon 
challenge \cite{rebuffi2017learning}, composed of ten different datasets 
of small images ($72\times 72$). Models are trained with a combination of
training and validation set and results obtained online. 
For fair comparison, we use the learning protocol of
\cite{rebuffi2017learning}. CovNorm achieves state of the art performance 
in terms of classification accuracy, parameter size, and decathlon score
${\bf S}$.

\vspace{-.1in}
\section{Conclusion}

CovNorm is an MDL technique of very simple implementation.
When compared to previous methods,
it dramatically reduces the number of adaptation parameters without 
loss of recognition performance. It was used to show
that large CNNs can be ``recycled'' across problems as
diverse as digit, object, scene, or fine-grained classes, with no
loss, by simply tuning $0.5\%$ of their
parameters. 

% This opens up exciting possibilities for adaptive
% applications, such as object tracking, or when
% little data is available, such as low-shot learning. We plan to
% investigate the use of CovNorm in these contexts in the future.

\vspace{-.1in}
\section{Acknowledgment}
This work was partially funded by NSF awards IIS-1546305 and IIS-1637941, a 
GRO grant from Samsung, and NVIDIA GPU donations.

{\small
	\bibliographystyle{ieee}
	\bibliography{egbib}
}

\end{document}